\pdfoutput=1

\documentclass[11pt]{article}

\def\numNERtasks{eight}
\def\numSENtasks{four}
\def\nummodels{14}
\def\numpretrainedscimodels{five}
\def\numtasks{12}

\usepackage[]{acl}

\usepackage{url,comment,amsmath,booktabs,graphicx,listings,multirow} 
\usepackage{times}
\usepackage{array}
\usepackage{latexsym}
\usepackage{siunitx}
\usepackage{xstring}
\usepackage[figuresleft]{rotating}
\usepackage{enumitem}
\usepackage{float}
\usepackage{balance}

\usepackage{adjustbox}
\usepackage{array}



\newcolumntype{R}[2]{%
    >{\adjustbox{angle=#1,lap=\width-(#2)}\bgroup}%
    l%
    <{\egroup}%
}

\usepackage[T1]{fontenc}

\usepackage[utf8]{inputenc}

\usepackage{microtype}

\usepackage{booktabs}

\usepackage{highlight}

\newif\iffinal

\newif\ifarxiv
\arxivfalse

\iffinal
    \newcommand\ben[1]{}
    \newcommand\ian[1]{}
    \newcommand\status[1]{}
    \newcommand\note[1]{}
    \newcommand\hong[1]{}
    \newcommand\kyle[1]{}
    \newcommand\eamon[1]{}
    \newcommand\aswathy[1]{}
    \newcommand\greg[1]{}
    \newcommand\carl[1]{}
\else
    \newcommand\ben[1]{{\color{blue}[Ben: #1]}}
    \newcommand\ian[1]{{\color{red}[Ian: #1]}}
    \newcommand\status[1]{{\color{purple}[Status \today{}: #1]}}
    \newcommand\note[1]{{\color{purple}[Note: #1]}}
    \newcommand\hong[1]{{\color{cyan}[Hong: #1]}}
    \newcommand\kyle[1]{{\color{green}[Kyle: #1]}}
    \newcommand\eamon[1]{{\color{violet}[Eamon: #1]}}
    \newcommand\aswathy[1]{{\color{orange}[Aswathy: #1]}}
    \newcommand\greg[1]{{\color{magenta}[Greg: #1]}}
    \newcommand\carl[1]{{\color{green}[Carl: #1]}}
    
\fi

\newcommand\SB{SB}

\title{The Diminishing Returns of Masked Language Models to Science }

\author{Zhi Hong$^*$, Aswathy Ajith$^*$, Gregory Pauloski$^*$, Eamon Duede$^{\dagger}$, \\  \textbf{Kyle Chard$^{*\ddag}$, Ian Foster$^{*\ddag}$} \\
        $^*$Department of Computer Science, University of Chicago, Chicago, IL 60637, USA \\
        $^\dagger$Department of Philosophy and Committee on Conceptual and Historical Studies of Science,\\ University of Chicago, Chicago, IL 60637, USA \\
        $^\natural$Public.Resource.Org, Healdsburg, CA 95448, USA \\
        $^\ddag$Data Science and Learning Division, Argonne National Laboratory, Lemont, IL 60615, USA
}

\begin{document}
\maketitle
\begin{abstract}

Transformer-based masked language models such as BERT, 
trained on general corpora, have shown impressive performance on downstream tasks. 
It has also been demonstrated that the downstream task performance of such models can be improved by pretraining larger models for longer on more data.
In this work, we empirically evaluate the extent to which these results extend to tasks in science.
We use 14 domain-specific transformer-based models (including \textsc{ScholarBERT}, a new 770M-parameter science-focused masked language model pretrained on up to 225B tokens) to evaluate the impact of training data, model size, pretraining and finetuning time on 12 downstream scientific tasks.
Interestingly, we find that increasing model sizes, training data, or compute time does not always lead to significant improvements (i.e., $>1\%$ F1), if at all, in scientific information extraction tasks and offered possible explanations for the surprising performance differences.

\end{abstract}


\section{Introduction}

Massive growth in the number of scientific publications places considerable cognitive burden on researchers \cite{teplitskiy2022status}. Language models can potentially serve as a tool to alleviate this burden by automating the scientific knowledge extraction process. BERT \cite{devlin2018bert} was pretrained on a general corpus (BooksCorpus and Wikipedia) which differs from scientific literature in terms of the context, terminology, and writing style~\cite{ahmad2012stylistic}. Subsequently, other masked language models have since been pretrained on domain-specific scientific corpora~\cite{gu2021domain,huang2022batterybert, beltagy2019scibert} with the goal of improving downstream task performance. 
(Here, we use the term \emph{domain} to indicate a specific scientific discipline such as biomedical science or computer science.)
Other studies~\cite{liu2019roberta, kaplan2020scaling} explored the impact of varying model size, training corpus size, and compute time on downstream task performance. However, no previous work has investigated how these parameters affect science-focused models.

In this study, we train a series of scientific language models, called \textsc{ScholarBERT}, on a large, multidisciplinary scientific corpus consisting of 225B tokens to understand the effects of model size, data size, as well as pretraining and finetuning epochs on downstream task performance. 
We find that for information extraction tasks, the primary application for scientific language models, the performance gains by training a larger model for longer with more data are not robust---they are highly dependent on the individual tasks.
We make the \textsc{ScholarBERT} models and a sample of the training corpus publicly available to encourage further studies.

\section{Related Work}

Prior research~\cite{kaplan2020scaling, brown2020language, liu2019roberta} has explored the effects of varying model size, dataset size, and amount of compute on language model performance.

\citet{kaplan2020scaling} demonstrated that cross-entropy training loss scales as a power-law with model size, dataset size, and compute time for unidirectional decoder-only architectures. \citet{brown2020language} showed that the few-shot learning abilities of language models can be improved by using larger models. However, both studies explored only the Generative Pre-trained Transformer (GPT), an autoregressive generative model~\cite{brown2020language}. 

By comparing BERT-Base (110M parameters) and BERT-Large (340M parameters), \citet{devlin2018bert} showed that masked language models can also benefit from larger models.
Likewise, the RoBERTa~\cite{liu2019roberta} paper demonstrates how BERT models can benefit from being trained for longer periods, with bigger batches, and with more data.



Models such as BERT and RoBERTa were pretrained on general corpora. To boost performance on scientific downstream tasks, SciBERT~\cite{beltagy2019scibert}, PubMedBERT~\cite{gu2021domain}, BioBERT~\cite{lee2020biobert}, and MatBERT~\cite{TREWARTHA2022100488} were trained on domain-specific text with the goal of enhancing performance on tasks requiring domain knowledge. Yet, as mentioned earlier, there is no work on how that task performance varies with pre-training parameters.

\section{Data and Methodology}
We outline the pretraining dataset, related models to which we compare performance, and the architecture and pretraining process used for creating the \textsc{ScholarBERT} models.

\subsection{The Public Resource Dataset}

We pretrain the \textsc{ScholarBERT} models on a dataset provided by Public.Resource.Org, Inc. (``Public Resource''), a nonprofit organization based in California.
This dataset was constructed from a corpus of 85M journal article PDF files, from which the Grobid tool, version 0.5.5, was used to extract text 
\cite{GROBID}. 
Not all extractions were successful, because of corrupted or badly encoded PDF files.
We work here with text from $\sim$75M articles in this dataset,
categorized as 45.3\% biomedicine, 23.1\% technology, 20.0\% physical sciences, 8.4\% social sciences, and 3.1\% arts \& humanities.
(A sample of the extracted texts and corresponding original PDFs is available in the Data attachment for review purposes.)

\begin{table*}
\centering
\begin{adjustbox}{width=\linewidth}
\begin{tabular}{|l|l|l|l|l|l|l|}
\hline
\textbf{Model} & \textbf{Architecture} & \textbf{Pretraining Method} & \textbf{Casing} & \textbf{Pretraining Corpus} &  \textbf{Domain} &  \textbf{Tokens} \\
\hline\hline
\texttt{BERT\_Base} & BERT-Base  & BERT & Cased & \texttt{Wiki} + \texttt{Books} & Gen & 3.3B \\
\texttt{SciBERT} & BERT-Base & BERT & Cased & \texttt{SemSchol} & Bio, CS & 3.1B \\
\texttt{PubMedBERT} & BERT-Base  & BERT & Uncased & \texttt{PubMed\textsubscript{A}} + \texttt{PMC} & Bio & 16.8B\\
\texttt{BioBERT\_1.2} & BERT-Base & BERT & Cased & \texttt{PubMed\textsubscript{B}} + \texttt{Wiki} + \texttt{Books} & Bio, Gen & 7.8B\\
\texttt{MatBERT} & BERT-Base & BERT & Cased & \texttt{MatSci} & Mat & 8.8B\\
\texttt{BatteryBERT} & BERT-Base & BERT & Cased & \texttt{Battery} & Mat & 5.2B\\
\hline
\texttt{BERT\_Large} & BERT-Large & BERT & Cased & \texttt{Wiki} + \texttt{Books} & Gen & 3.3B \\
\texttt{ScholarBERT\_1} & BERT-Large & RoBERTa-like & Cased & \texttt{PRD\_1} & Sci & 2.2B\\
\texttt{ScholarBERT\_10} & BERT-Large & RoBERTa-like & Cased & \texttt{PRD\_10} & Sci & 22B \\
\texttt{ScholarBERT\_100} & BERT-Large & RoBERTa-like & Cased & \texttt{PRD\_100} & Sci & 221B \\
\texttt{ScholarBERT\_10\_WB} & BERT-Large & RoBERTa-like & Cased & \texttt{PRD\_10} + \texttt{Wiki} + \texttt{Books} & Sci, Gen & 25.3B\\
\texttt{ScholarBERT\_100\_WB} & BERT-Large & RoBERTa-like & Cased &  \texttt{PRD\_100} + \texttt{Wiki} + \texttt{Books} & Sci, Gen & 224.3B\\
\hline
\texttt{ScholarBERT-XL\_1} & BERT-XL & RoBERTa-like & Cased & \texttt{PRD\_1}  & Sci & 2.2B\\
\texttt{ScholarBERT-XL\_100} & BERT-XL & RoBERTa-like & Cased & \texttt{PRD\_100} & Sci & 221B\\
\hline
\end{tabular}
\end{adjustbox}
\caption{
Characteristics of the \nummodels{} BERT models considered in this study.
The BERT-Base and -Large architectures are described in~\cite{devlin2018bert};
the BERT-XL architecture has 36 layers, hidden size of 1280, and 20 heads.
Details of the pretraining corpora are in \autoref{tab:pretraining-datasets} in the Appendix.
The domains are Bio=biomedicine, CS=computer science, Gen=general, Mat=materials science and engineering, and Sci=broad scientific.
}
\label{tab:models}
\end{table*}
\subsection{Models}

We consider \nummodels{} BERT models: seven from existing literature (BERT-Base, BERT-Large, SciBERT, PubMedBERT, BioBERT v1.2, MatBERT, and BatteryBERT: \autoref{sec:appendix-extant-bert}); and seven \textsc{ScholarBERT} variants pretrained on different subsets of the Public Resource dataset (and, in some cases, also the WikiBooks corpus).
We distinguish these models along the four dimensions listed in \autoref{tab:models}: 
architecture, pretraining method, pretraining corpus, and casing.
\textsc{ScholarBERT} and \textsc{ScholarBERT-XL}, with 340M and 770M parameters, respectively, are the largest science-specific BERT models reported to date.
Prior literature demonstrates the efficacy of pretraining BERT models on domain-specific corpora~\cite{sun2019fine,fabien2020bertaa}. However, 
the ever-larger scientific literature 
makes pretraining domain-specific language models
prohibitively expensive. A promising alternative is to create larger, multi-disciplinary BERT models, such as \textsc{ScholarBERT}, that harness the increased availability of diverse pretraining text; researchers can then adapt (i.e., finetune) these general-purpose science models to meet their specific needs. 



\subsection{\textsc{ScholarBERT} Pretraining}
\label{sec:pro-pretraining}




We randomly sample 1\%, 10\%, and 100\% of the Public Resource dataset to create PRD\_1, PRD\_10, and PRD\_100.
We pretrain \textsc{ScholarBERT} models on these PRD subsets by using the RoBERTa pretraining procedure, which has been shown to produce better downstream task performance in a variety of domains~\cite{liu2019roberta}.
See Appendix~\ref{sec:roberta-opt} for details.

\section{Experimental Results}
\label{sec:experimental_results}

We first perform sensitivity analysis across ScholarBERT pretraining dimensions to 
determine the trade-off between time spent in pretraining versus finetuning.
We also compare the downstream task performance of \textsc{ScholarBERT} to that achieved with other BERT models.
Details of each evaluation task are in Appendix~\ref{sec:appendix-eval-tasks}.

\subsection{Sensitivity Analysis}
\label{sec:sensitivity}

We save checkpoints periodically while pretraining each \textsc{ScholarBERT(-XL)} model. In this analysis, we select the checkpoints 
at $\sim$0.9k, 5k, 10k, 23k, and 33k iterations based on the decrease of training loss between iterations.
We observe that pretraining loss decreases rapidly until around \num{10000} iterations;
further training to convergence (roughly \num{33000} iterations) yields small decreases of training loss: see \autoref{fig:train-loss} in Appendix.

To measure how downstream task performance is impacted by pre-training and finetuning time, we finetune each of the checkpointed models for 5 and 75 epochs. We observe the following: 
(1) The under-trained 0.9k-iteration model sees the biggest boost in the F1 scores of downstream tasks (+8\%) with more finetuning, but even with 75 epochs of finetuning the 0.9k-iteration models' average F1 score is still 19.9 percentage points less than that of the 33k-iteration model with 5 epochs of finetuning.
(2) For the subsequent checkpoints, the performance gains from more finetuning decreases as the number of pre-training iterations increases. The average downstream task performance of the 33k-iteration model is only 0.39 percentage points higher with 75 epochs of finetuing than with 5 epochs.
Therefore, in the remaining experiments, we use the \textsc{ScholarBERT}(-XL) model that was pretrained for 33k iterations and  finetuned for 5 epochs.

\begin{table*}[ht!]
\centering
\begin{small}
\resizebox{\textwidth}{!}{%
\begin{tabular}
{|r|I{1.3cm}I{1.3cm}I{2cm}I{1.5cm}|I{1.8cm}|I{1.8cm}|I{1.8cm}|I{1.5cm}|I{0.8cm}|}
\hline
    \multicolumn{1}{|r|}{\textbf{Domain}} &
    \multicolumn{4}{c|}{\textbf{Biomedical}} &
    \multicolumn{1}{c|}{\textbf{CS}} &
    \multicolumn{1}{c|}{\textbf{Materials}} &
    \multicolumn{1}{c|}{\textbf{Multi-Domain}} &
    \multicolumn{1}{c|}{\textbf{Sociology}} &
    \multicolumn{1}{c|}{} \\
\cline{1-9}
    \textbf{Dataset} &
    \multicolumn{1}{c}{\textbf{BC5CDR}} &
    \multicolumn{1}{c}{\textbf{JNLPBA}} &
    \multicolumn{1}{c}{\textbf{NCBI-Disease}} &
    \multicolumn{1}{c|}{\textbf{ChemDNER}} &
    \multicolumn{1}{c|}{\textbf{SciERC}} &
    \multicolumn{1}{c|}{\textbf{MatSciNER}} &
    \multicolumn{1}{c|}{\textbf{ScienceExam}} &
    \multicolumn{1}{c|}{\textbf{Coleridge}} &
    \multicolumn{1}{c|}{\textbf{Mean}} \\
\hline\hline
\texttt{BERT-Base} & 85.36 & 72.15 & 84.28 & 84.84 & 56.73 & 78.51 & 78.37 & 57.75 & 74.75 \\
\texttt{BERT-Large} & 86.86 & 72.80 & 84.91 & 85.83 & 59.20 & 82.16 & 82.32 & 57.46 & 76.44 \\
\texttt{SciBERT} & 88.43 & 73.24 & 86.95 & 85.76 & +\underline{59.36}- & 82.64 & 78.83 & 54.07 & 76.16 \\
\texttt{PubMedBERT} & +\underline{89.34}- & +\underline{74.53}- & +\underline{87.91}- & +\underline{87.96}- & 59.03 & 82.63 & 69.73 & 57.71 & 76.11 \\
\texttt{BioBERT} & \underline{88.01} & \underline{73.09} & \underline{87.84} & \underline{85.53} & 58.24 & 81.76 & 78.60 & 57.04 & 76.26 \\
\texttt{MatBERT} & 86.44 & 72.56 & 84.94 & 86.09 & 58.52 & +\underline{83.35}- & 80.01 & 56.91 & 76.10 \\
\texttt{BatteryBERT} & 87.42 & 72.78 & 87.04 & 86.49 & 59.00 & \underline{82.94} & 78.14 & +59.87- & +76.71- \\
\hline
\texttt{\SB{}\_1} & 87.27 & 73.06 & 85.49 & 85.25 & 58.62 & 80.87 & \underline{82.75} & 55.34 & 76.08 \\
\texttt{\SB{}\_10} & 87.69 & 73.03 & 85.65 & 85.80 & 58.39 & 80.61 & +\underline{83.24}- & 53.41 & 75.98 \\
\texttt{\SB{}\_100} & 87.84 & 73.47 & 85.92 & 85.90 & 58.37 & 82.09 & \underline{83.12} & 54.93 & 76.46 \\
\texttt{\SB{}\_10\_WB} & 86.68 & 72.67 & 84.51 & 83.94 & 57.34 & 78.98 & \underline{83.00} & 54.29 & 75.18 \\
\texttt{\SB{}\_100\_WB} & 86.89 & 73.16 & 84.88 & 84.31 & 58.43 & 80.84 & \underline{82.43} & 54.00 & 75.62 \\
\texttt{\SB{}-XL\_1} & 87.09 & 73.14 & 84.61 & 85.81 & 58.45 & 82.84 & \underline{81.09} & 55.94 & 76.12 \\
\texttt{\SB{}-XL\_100} & 87.46 & 73.25 & 84.73 & 85.73 & 57.26 & 81.75 & \underline{80.72} & 54.54 & 75.68 \\
\hline
\end{tabular}
}
\caption{NER F1 scores for each model. Models are finetuned five times for each dataset and the average result is presented. 
Underlined results represent the F1-scores of models trained on in-distribution data for the given task, and bolded results indicate the best performing model on that task.
\texttt{SB} = \textsc{ScholarBERT}.
\label{tab:token-cls}
}
\end{small}
\end{table*}

\subsection{Finetuning}
\label{sec:token_cls}

We finetuned the \textsc{ScholarBERT} models and the state-of-the-art scientific models listed in \autoref{tab:models} on NER, relation extraction, and sentence classification tasks. F1 scores for each model-task pair, averaged over five runs, are shown in Tables~\ref{tab:token-cls} and~\ref{tab:sent-cls}.
For NER tasks, we use the CoNLL NER evaluation Perl script~\cite{conlleval} to compute F1 scores for each test.

\begin{table*}[ht!]
\centering
\resizebox{1.2\columnwidth}{!}{%
\begin{small}
\begin{tabular}{|r|I{1.2cm}|I{1.5cm}|I{1.5cm}|I{1.2cm}|I{0.8cm}|}
\hline
    \multicolumn{1}{|r|}{\textbf{Domain}} &
    \multicolumn{1}{c|}{\textbf{CS}} &
    \multicolumn{1}{c|}{\textbf{Biomedical}} &
    \multicolumn{1}{c|}{\textbf{Multi-Domain}} &
    \multicolumn{1}{c|}{\textbf{Materials}} &
    \multicolumn{1}{c|}{} \\
\cline{1-5}

{\textbf{Dataset}} & \multicolumn{1}{c|}{\textbf{SciERC}} & \multicolumn{1}{c|}{\textbf{ChemProt}} & \multicolumn{1}{c|}{\textbf{PaperField}} & \multicolumn{1}{c|}{\textbf{Battery}} & \multicolumn{1}{c|}{\textbf{Mean}} \\
\hline\hline
\texttt{BERT-Base} & 74.95 & 83.70 & 72.83 & 96.31 & 81.95 \\
\texttt{BERT-Large} & 80.14 & 88.06 & 73.12 & +96.90- & 84.56 \\
\texttt{SciBERT} & \underline{79.26} & 89.80 & 73.19 & 96.38 & 84.66 \\
\texttt{PubMedBERT} & 77.45 & +\underline{91.78}- & +73.93- & 96.58 & +84.94- \\
\texttt{BioBERT} & 80.12 & \underline{89.27} & 73.07 & 96.06 & 84.63 \\
\texttt{MatBERT} & 79.85 & 88.15 & 71.50 & \underline{96.33} & 83.96 \\
\texttt{BatteryBERT} & 78.14 & 88.33 & 73.28 & \underline{96.06} & 83.95 \\
\hline
\texttt{\SB{}\_1} & 73.01 & 83.04 & \underline{72.77} & 94.67 & 80.87 \\
\texttt{\SB{}\_10} & 75.95 & 82.92 & \underline{72.94} & 92.83 & 81.16 \\
\texttt{\SB{}\_100} & 76.19 & 87.60 & \underline{73.14} & 92.38 & 82.33 \\
\texttt{\SB{}\_10\_WB} & 73.17 & 81.48 & \underline{72.37} & 93.15 & 80.04\\
\texttt{\SB{}\_100\_WB} & 76.71 & 83.98 & \underline{72.29} & 95.55 & 82.13 \\
\texttt{\SB{}-XL\_1} & 74.85 & 90.60 & \underline{73.22} & 88.75 & 81.86 \\
\texttt{\SB{}-XL\_100} & +80.99- & 89.18 & \underline{73.66} & 95.44 & 84.82 \\
\hline
\end{tabular}
\end{small}
}
\caption{F1 scores for each model on Relation Extraction (SciERC, ChemProt) and Sentence Classification (PaperField, Battery) tasks. Models are finetuned five times for each dataset and the average result is presented.
Underlined results represent the F1-scores of models trained on in-distribution data for the given task, and bolded results indicate the best performing model on that task.
\texttt{SB} = \textsc{ScholarBERT}.
}
\label{tab:sent-cls}
\end{table*}

Tables~\ref{tab:token-cls} and~\ref{tab:sent-cls} show the results, from which we can make the following observations:
(1) With the same training data, a larger model cannot always achieve significant performance improvements. 
BERT-Base achieved F1 scores within 1 percentage point of BERT-Large on 6/\numtasks{} tasks;
SB\_1 achieved F1 scores within 1 percentage point of SB-XL\_1 on 7/\numtasks{} tasks;
SB\_100 achieved F1 scores within 1 percentage point of SB-XL\_100 on 6/\numtasks{} tasks.
(2) With the same model size, a model pretrained on more data cannot guarantee significant performance improvements.
SB\_1 achieved F1 scores within 1 percentage point of SB\_100 on 8/\numtasks{} tasks;
SB\_10\_WB achieved F1 scores within 1 percentage point of SB\_100\_WB on 7/\numtasks{} tasks;
SB-XL\_1 achieved F1 scores within 1 percentage point of SB-XL\_100 on 10/\numtasks{} tasks.
(3) Domain-specific pretraining cannot guarantee significant performance improvements. The Biomedical domain is the only domain where we see the 
on-domain model 
(i.e., pretrained for the associated domain; marked with underlines; in this case is PubMedBERT) consistently outperformed models pretrained on off-domain or more general corpora by more than 1 percentage point F1. The same cannot be said for  CS, Materials, or Multi-Domain tasks.

\subsection{Discussion}
Here we offer possible explanations for the three observations stated above.
(1) The nature of the task is more indicative of task performance than the size of the model. In particular, with the same training data, a larger model size impacts performance only for relation extraction tasks, which consistently saw F1 scores increase by more than 1 percentage point when going from smaller models to larger models (i.e., BERT-Base to BERT-Large, SB\_1 to SB-XL\_1, SB\_100 to SB-XL\_100). In contrast, the NER and sentence classification tasks did not see such consistent significant improvements.
(2) Our biggest model, \textsc{ScholarBERT-XL}, is only twice as large as the original BERT-Large, but its pretraining corpus is 100X larger. 
The training loss of the \textsc{ScholarBERT-XL\_100} model dropped rapidly only in the first $\sim$10k iterations (Fig.~\ref{fig:train-loss} in Appendix), which covered the first 1/3 of the PRD corpus, thus
it is possible that the PRD corpus can saturate even our biggest model.
~\cite{kaplan2020scaling, hoffmann2022training}.
(3) Finetuning can compensate for missing domain-specific knowledge in pretraining data. While pretraining language models on a specific domain can help learn domain-specific concepts, finetuning can also fill holes in the pretraining corpora's domain knowledge, as long as the pretraining corpus incorporates the characteristics specific to the finetuning dataset.

\section{Conclusions}

We have reported experiments that compare and evaluate the impact of various parameters (model size, pretraining dataset size and breadth, and pretraining and finetuning lengths) on the performance of different language models pretrained on scientific literature. Our results encompass \nummodels{} existing and newly-developed BERT-based language models across \numtasks{} scientific downstream tasks.


We find that model performance on downstream scientific information extraction tasks is not improved significantly or consistently by increasing \textit{any} of the four parameters considered (model size, amount of pretraining data, pretraining time, finetuning time).
We attribute these results to both the power of finetuning and limitations in the evaluation datasets, as well as (for the \textsc{ScholarBERT} models) small model sizes relative to the large pretraining corpus. 


\ifarxiv
\else
We will make all pretrained \textsc{ScholarBERT} models, plus a subset of the Public Resource Dataset, freely available online.
(We are not permitted to share the full Public Resource Dataset.)
\fi


\section*{Limitations}
Our 12 labeled test datasets are from just five domains (plus two multi-disciplinary); five of the 12 are from biomedicine. This imbalance, which reflects the varied adoption of NLP methods across domains, means that our evaluation dataset is necessarily limited.
Our largest model, with 770M parameters, may not be sufficiently large to demonstrate scaling laws for language models.
We also aim to extend our experiments to tasks other than NER, relation extraction, and text classification, such as question-answering and textual entailment in scientific domains.



\bibliography{anthology,custom}
\bibliographystyle{acl_natbib}

\newpage

\appendix

\section{Extant BERT-based models}
\label{sec:appendix-extant-bert}
\citet{devlin2018bert} introduced BERT-Base and BERT-Large, with $\sim$110M and $\sim$340M parameters, as transformer-based masked language models conditioned on both the left and right contexts. Both are pretrained on the English Wikipedia + BooksCorpus datasets.  

SciBERT~\cite{beltagy2019scibert} follows the BERT-Base architecture and is pretrained on data from two domains, namely,  biomedical science and computer science. SciBERT outperforms BERT-Base on finetuning tasks by an average of 1.66\% and 3.55\% on biomedical tasks and computer science tasks, respectively.

BioBERT~\cite{lee2020biobert} is a BERT-Base model with a pretraining corpus from PubMed abstracts and full-text PubMedCentral articles. 
Compared to BERT-Base, BioBERT achieves improvements of 0.62\%, 2.80\%, and 12.24\% on biomedical NER, biomedical relation extraction, and biomedical question answering, respectively.

PubMedBERT~\cite{gu2021domain}, another BERT-Base model targeting the biomedical domain, is also pretrained on PubMed and PubMedCentral text.
However, unlike BioBERT, PubMedBERT is trained as a new BERT-Base model, using text drawn exclusively from PubMed and PubMedCentral. As a result, the vocabulary used in PubMedBERT varies significantly from that used in BERT and BioBERT. Its pretraining corpus contains 3.1B words from PubMed abstracts and 13.7B words from PubMedCentral articles.
PubMedBERT achieves state-of-the-art performance on the Biomedical Language Understanding and Reasoning Benchmark, outperforming BERT-Base by 1.16\%~\cite{gu2021domain}. 

MatBERT~\cite{TREWARTHA2022100488} is a materials science-specific model pretrained on 2M journal articles (8.8B tokens). It consistently outperforms BERT-Base and SciBERT in recognizing materials science entities related to solid states, doped materials, and gold nanoparticles, with $\sim$10\% increase in F1 score compared to BERT-Base, and a 1\% to 2\% improvement compared to SciBERT.

BatteryBERT~\cite{huang2022batterybert} is a model pretrained on \num{400366} battery-related publications (5.2B tokens). BatteryBERT has been shown to outperform BERT-Base by less than 1\% on the SQuAD  question answering task. For battery-specific question-answering tasks, its F1 score is around 5\% higher than that of BERT-base.

\section{ScholarBERT Pretraining Details}

\begin{figure*}[ht]
\begin{center}
    \includegraphics[width=\textwidth, trim=2mm 2mm 2mm 2mm, clip]{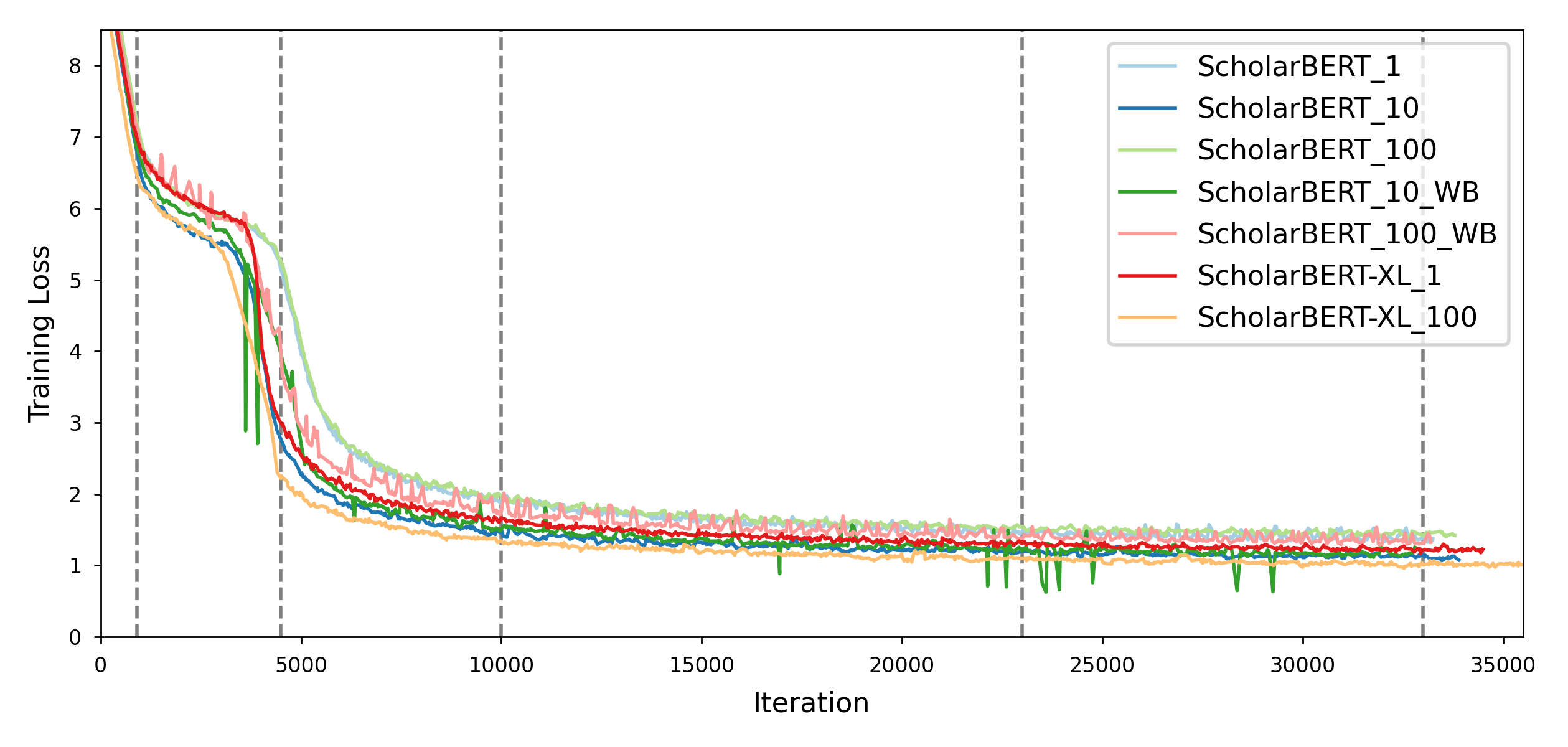}
    
    \vspace{-1ex}
    
    \caption{
    Pretraining loss plots for the \textsc{ScholarBERT} models listed in \autoref{tab:models}.
    The vertical dashed lines indicate the approximate locations of the iteration checkpoints selected for evaluation in \autoref{sec:sensitivity}.
    \label{fig:train-loss}}
\end{center}
\end{figure*}

\begin{table*}[ht!]
\centering
\begin{small}
\begin{tabular}{|l|p{10.5cm}|c|c|}
\hline
\textbf{Name} & \textbf{Description} &  \textbf{Domain} & \textbf{Tokens} \\
\hline\hline
\texttt{Wiki} & English-language Wikipedia articles~\cite{wikicorpus} & Gen & 2.5B\\
\hline\texttt{Books} & BookCorpus~\cite{zhu2015aligning,wikicorpus}: Full text of 11038 books  & Gen & 0.8B \\
\hline
\texttt{SemSchol} & 1.14M papers 
from Semantic Scholar~\cite{cohan2019structural}, 18\% in CS, 82\% in Bio & Bio, CS & 3.1B \\
\hline
\texttt{PubMed\textsubscript{A}} & Biomedical abstracts sampled from PubMed~\cite{gu2021domain} & Bio &  3.1B \\
\hline
\texttt{PubMed\textsubscript{B}} & Biomedical abstracts sampled from PubMed~\cite{lee2020biobert} & Bio &  4.5B \\
\hline
\texttt{PMC} & Full-text biomedical articles sampled from PubMedCentral~\cite{gu2021domain} & Bio & 13.7B \\
\hline
\texttt{MatSci} & 2M peer-reviewed materials science journal articles~\cite{TREWARTHA2022100488} & Materials & 8.8B \\
\hline
\texttt{Battery} & \num{0.4}M battery-related publications~\cite{huang2022batterybert} & Materials & 5.2B \\
\hline
\texttt{PRD\_1} & 1\% of the English-language research articles from the Public Resource dataset& Sci & 2.2B \\
\hline
\texttt{PRD\_10} & 10\% of the English-language research articles from the Public Resource dataset& Sci & 22B \\
\hline
\texttt{PRD\_100} & 100\% of the English-language research articles from the Public Resource dataset& Sci & 221B \\
\hline
\end{tabular}
\end{small}

\vspace{-1ex}

\caption{
Pretraining corpora used by models in this study.
The domains are Bio=biomedicine, CS=computer science, Gen=general, Materials=materials science and engineering and Sci=broad scientific.}
\label{tab:pretraining-datasets}
\end{table*}

\subsection{Tokenization}

The vocabularies generated for PRD\_1 and PRD\_10 differed only in 1--2\% of the tokens; however, in an initial study, the PRD\_100 vocabulary differed from that of PRD\_10 by 15\%.
A manual inspection of the PRD\_100 vocabulary revealed that many common English words such as ``is,'' ``for,''  and ``the'' were missing. We determined that these omissions were an artifact of PRD\_100 being sufficiently large to cause integer overflows in the unsigned 32-bit-integer token frequency counts used by HuggingFace's tokenizers library.
For example, ``the'' was not in the final vocabulary because the token ``th'' overflowed. Because WordPiece iteratively merges smaller tokens to create larger ones, the absence of tokens like ``th'' or ``\#\#he'' means that ``the'' could not appear in the final vocabulary.

We modified the tokenizers library to use unsigned 64-bit integers for all frequency counts, 
and recreated a correct vocabulary for PRD\_100.
Interestingly, models trained on the PRD\_100 subset with the incorrect and correct vocabularies 
exhibited comparable performance on downstream tasks. 

\subsection{RoBERTa Optimizations}
\label{sec:roberta-opt}

RoBERTa introduces many optimizations for improving BERT pretraining performance~\cite{liu2019roberta}.
1) It uses a single phase training approach whereby all training is performed with a maximum sequence length of 512.
2) Unlike BERT which randomly introduces a small percentage of shortened sequence lengths into the training data, RoBERTa does not randomly use shortened sequences.
3) RoBERTa uses dynamic masking, meaning that each time a batch of training samples is selected at runtime, a new random set of masked tokens is selected;
in contrast, BERT uses static masking, pre-masking the training samples prior to training. BERT duplicates the training data 10 times each with a different random, static masking.
4) RoBERTa does not perform Next Sentence Prediction  during training.
5) RoBERTa takes sentences contiguously from one or more documents until the maximum sequence length is met.
6) RoBERTa uses a larger batch size of \num{8192}.
7) RoBERTa uses byte-pair encoding (BPE) rather than WordPiece.
8) RoBERTa uses an increased vocabulary size  of \num{50000}, 67\% larger than BERT. 
9) RoBERTa trains for more iterations (up to \num{500000}) than does BERT-Base (\num{31000}).

We adopt RoBERTa training methods, with three key exceptions.
1) Unlike RoBERTa, we randomly introduce smaller length samples because many of our downstream tasks use sequence lengths much smaller than the maximum sequence length of 512 that we pretrain with. 2) We pack training samples with sentences drawn from a single document, as the RoBERTa authors note that this results in slightly better performance. 3) We use WordPiece encoding rather than BPE, as the RoBERTa authors note that BPE can result in slightly worse downstream performance.

\subsection{Hardware and Software Stack}
\label{sec:appendix-hardware-software}

\begin{table}[ht]
\centering
\begin{small}
\begin{tabular}{|r|l|}
\hline
\textbf{Hyperparameter} & \textbf{Value} \\
\hline\hline
Steps & \num{33000} \\
Optimizer & LAMB \\
LR & 0.0004 \\
LR Decay & Linear \\
LR Warmup Steps & 0.06\% \\
Batch Size & \num{32768} \\
Precision & FP16 \\
Weight Decay & 0.01 \\
Attention Dropout & 10\% \\
Hidden Dropout & 10\% \\
Hidden Activation & GELU \\
\hline
\end{tabular}
\end{small}
\caption{
    Pretraining hyperparameters. All \textsc{ScholarBERT} variants use the same pretraining hyperparameters.
    \label{tab:hyperparams}
}
\end{table}

We perform data-parallel pretraining on a cluster with 24 nodes, each containing eight 40~GB NVIDIA A100 GPUs.
In data-parallel distributed training, a copy of the model is replicated on each GPU, and, in each iteration, each GPU computes on a unique local mini-batch.
At the end of the iteration, the local gradients of each model replica are averaged to keep each model replica in sync.
We perform data-parallel training of \textsc{ScholarBERT} models using PyTorch's distributed data-parallel model wrapper and 16 A100 GPUs.
For the larger \textsc{ScholarBERT-XL} models, we use the DeepSpeed data-parallel model wrapper and 32 A100 GPUs.
The DeepSpeed library incorporates a number of optimizations that improve training time and reduced memory usage, enabling us to train the larger model in roughly the same amount of time as the smaller model. 

We perform training in FP16 with a batch size of \num{32768} for $\sim$\num{33000} iterations (\autoref{tab:hyperparams}).
To achieve training with larger batch sizes, we employ NVIDIA Apex's FusedLAMB~\cite{nvidiaapex} optimizer, with an initial learning rate of 0.0004.
The learning rate is warmed up for the first 6\% of iterations and then linearly decayed for the remaining iterations.
We use the same masked token percentages as are used for BERT.
Training each model requires roughly 1000 node-hours, or 8000 GPU-hours.

\autoref{fig:train-loss} depicts the pretraining loss for each \textsc{ScholarBERT} model. We train each model past the point of convergence and take checkpoints throughout training to evaluate model performance as a function of training time.

\section{Evaluation Tasks}
\label{sec:appendix-eval-tasks}

We evaluate the models on \numNERtasks{} NER tasks and \numSENtasks{} sentence-level tasks.
For the NER tasks, we use 
\numNERtasks{} 
annotated scientific NER datasets:

\noindent
\begin{enumerate}
    \item BC5CDR~\cite{li2016biocreative}: An NER dataset identifying diseases, chemicals, and their interactions, generated from the abstracts of \num{1500} PubMed articles containing \num{4409} annotated chemicals, \num{5818} diseases, and \num{3116} chemical-disease interactions, totaling \num{6283} unique entities.
    \item JNLPBA~\cite{kim2004introduction}: A bio-entity recognition dataset of molecular biology concepts from \num{2404} MEDLINE abstracts, consisting of \num{21800} unique entities.
    \item SciERC~\cite{luan2018multi}: A dataset annotating entities, relations, and coreference clusters in 500 abstracts from 12 AI conference/workshop proceedings. It contains \num{5714} distinct named entities. 
    \item NCBI-Disease~\cite{dougan2014ncbi}: Annotations for 793 PubMed abstracts: \num{6893} disease mentions, of which \num{2134} are unique.
    \item ChemDNER~\cite{krallinger2015chemdner}: A chemical entity recognition dataset derived from \num{10000} abstracts containing \num{19980} unique chemical entity mentions.
    \item MatSciNER~\cite{TREWARTHA2022100488}: 800 annotated abstracts from solid state materials publications sourced via Elsevier's Scopus/ScienceDirect, Springer-Nature, Royal Society of Chemistry, and Electrochemical Society.
    Seven types of entities are labeled: inorganic materials (MAT), symmetry/phase labels (SPL), sample descriptors (DSC), material properties (PRO), material applications (APL), synthesis methods (SMT), and characterization methods (CMT).
    \item ScienceExam~\cite{smith2019scienceexamcer}: 133K entities from the Aristo Reasoning Challenge Corpus of 3rd to 9th grade science exam questions.
    \item Coleridge~\cite{staff2020coleridge}: \num{13588} entities from sociology articles indexed by the Inter-university Consortium for Political and Social Research (ICPSR).
\end{enumerate}

The sentence-level downstream tasks are relation extraction on the ChemProt (biology) and SciERC (computer science) datasets, and sentence classification on the Paper Field (multidisciplinary) and Battery (materials) dataset:
\begin{enumerate}
    \item
ChemProt consists of 1820 PubMed abstracts with chemical-protein interactions annotated by domain experts~\cite{peng2019transfer}.
\item 
SciERC, introduced above, provides \num{4716} relations \cite{luan2018multi}.
\item 
The Paper Field dataset~\cite{beltagy2019scibert},
built from the Microsoft Academic Graph~\cite{sinha2015overview}, maps paper titles to one of seven fields of study (geography, politics, economics, business, sociology, medicine, and psychology), with each field of study having around 12K training examples.
\item The Battery Document Classification dataset~\cite{huang2022batterybert} includes \num{46663} paper abstracts, of which \num{29472} are labeled as battery and the other \num{17191} as non-battery. The labeling is performed in a semi-automated manner. Abstracts are selected from 14 battery journals and \num{1044} non-battery journals, with the former labeled ``battery'' and the latter ``non-battery.''
\end{enumerate}

\section{Extended Results}
\label{sec:appendix-ext-results}

\begin{figure*}[!ht]
    \centering
    \includegraphics[width=\textwidth,trim=3mm 3mm 2mm 3mm,clip]{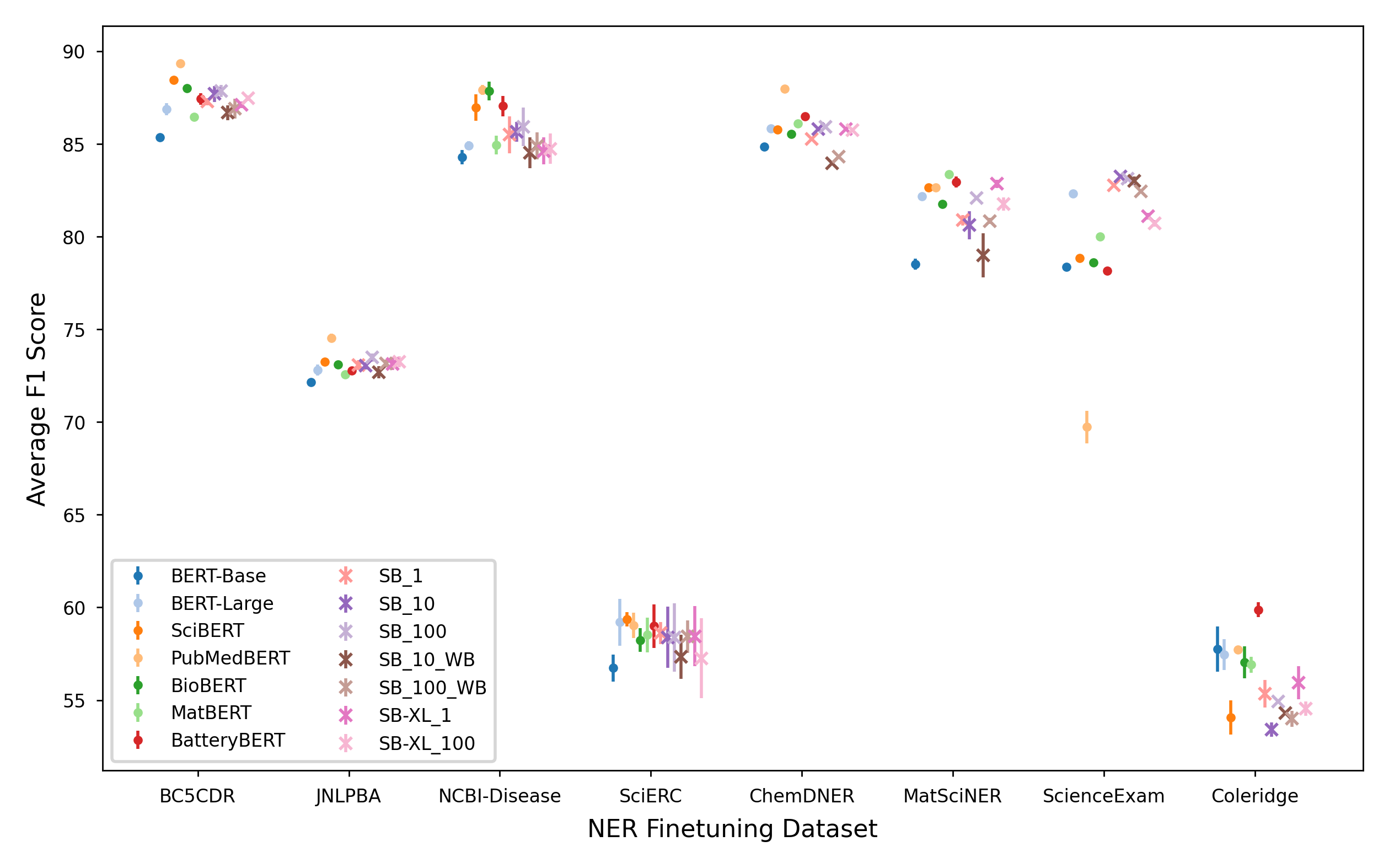}
    \caption{NER F1 scores from \autoref{tab:token-cls-appendix}, with standard deviations represented by error bars.}
    \label{fig:ner-error-bars}
\end{figure*}

\begin{figure*}[!ht]
    \centering
    \includegraphics[width=\textwidth]{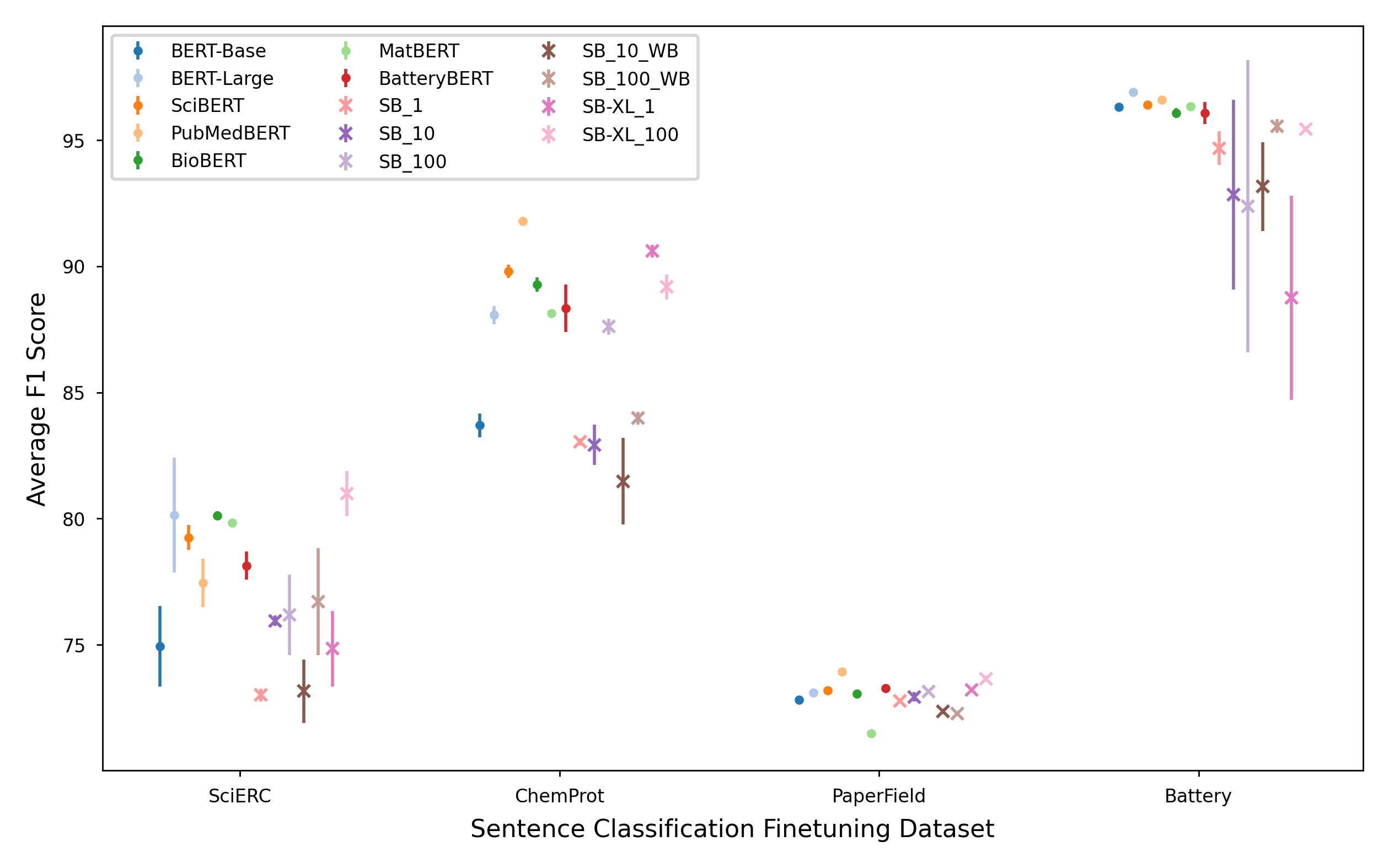}
    
    \vspace{-1ex}
    
    \caption{Sentence classification F1 scores from \autoref{tab:sent-cls-appendix}, with standard deviations represented by error bars.} 
    \label{fig:sent-cls-error-bars}
\end{figure*}

\autoref{tab:token-cls-appendix} shows average F1 scores with standard deviations for the NER tasks, each computed over five runs; \autoref{fig:ner-error-bars} presents the same data, with standard deviations represented by error bars.
\autoref{tab:sent-cls-appendix} and \autoref{fig:sent-cls-error-bars} show the same for sentence classification tasks. The significant overlaps of error bars for NCBI-Disease, SciERC NER, Coleridge, SciERC Sentence Classification, and ChemProt corroborate our observation in \autoref{sec:experimental_results} that on-domain pretraining provides only marginal advantage for downstream prediction over pretraining on a different domain or a general corpus.

\begin{table*}[ht!]
\centering
\begin{small}
\begin{tabular}{|l|cccc|}
\hline
{} & \textbf{BC5CDR} & \textbf{JNLPBA} & \textbf{NCBI-Disease} & \textbf{SciERC} \\
\hline\hline
\texttt{BERT-Base} & $85.36 \pm 0.189$ & $72.15 \pm 0.118$ & $84.28 \pm 0.388$ & $56.73 \pm 0.716$  \\
\texttt{BERT-Large} & $86.86 \pm 0.321$ & $72.80 \pm 0.299$ & $84.91 \pm 0.229$ & $59.20 \pm 1.260$  \\
\texttt{SciBERT} & $88.43 \pm 0.112$ & $73.24 \pm 0.184$ & $86.95 \pm 0.714$ & $59.36 \pm 0.390$  \\
\texttt{PubMedBERT} & $89.34 \pm 0.185$ & $74.53 \pm 0.220$ & $87.91 \pm 0.267$ & $59.03 \pm 0.688$ \\
\texttt{BioBERT} & $88.01 \pm 0.133$ & $73.09 \pm 0.230$ & $87.84 \pm 0.513$ & $58.24 \pm 0.631$ \\
\texttt{MatBERT} & $86.44 \pm 0.156$ & $72.56 \pm 0.162$ & $84.94 \pm 0.504$ & $58.52 \pm 0.933$ \\
\texttt{BatteryBERT} & $87.42 \pm 0.308$ & $72.78 \pm 0.190$ & $87.04 \pm 0.553$ & $59.00 \pm 1.174$ \\
\hline
\texttt{\SB{}\_1} & $87.27 \pm 0.189$ & $73.06 \pm 0.265$ & $85.49 \pm 0.998$ & $58.62 \pm 0.602$ \\
\texttt{\SB{}\_10} & $87.69 \pm 0.433$ & $73.03 \pm 0.187$ & $85.65 \pm 0.544$ & $58.39 \pm 1.643$ \\
\texttt{\SB{}\_100} & $87.84 \pm 0.329$ & $73.47 \pm 0.210$ & $85.92 \pm 1.040$ & $58.37 \pm 1.845$ \\
\texttt{\SB{}\_10\_WB} & $86.68 \pm 0.397$ & $72.67 \pm 0.329$ & $84.51 \pm 0.838$ & $57.34 \pm 1.199$ \\
\texttt{\SB{}\_100\_WB} & $86.89 \pm 0.543$ & $73.16 \pm 0.211$ & $84.88 \pm 0.729$ & $58.43 \pm 0.881$ \\
\texttt{\SB{}-XL\_1} & $87.09 \pm 0.179$ & $73.14 \pm 0.352$ & $84.61 \pm 0.730$ & $58.45 \pm 1.614$ \\
\texttt{\SB{}-XL\_100} & $87.46 \pm 0.142$ & $73.25 \pm 0.300$ & $84.73 \pm 0.817$ & $57.26 \pm 2.146$ \\
\hline
\hline
{} & \textbf{ChemDNER} & \textbf{MatSciNER} & \textbf{ScienceExam} & \textbf{Coleridge} \\
\hline\hline
\texttt{BERT-Base} & $84.84 \pm 0.004$ & $78.51 \pm 0.300$ & $78.37 \pm 0.004$ & $57.75 \pm 1.230$ \\
\texttt{BERT-Large} & $85.83 \pm 0.022$ & $82.16 \pm 0.040$ & $82.32 \pm 0.072$ & $57.46 \pm 0.818$ \\
\texttt{SciBERT} & $85.76 \pm 0.089$ & $82.64 \pm 0.054$ & $78.83 \pm 0.004$ & $54.07 \pm 0.930$ \\
\texttt{PubMedBERT} & $87.96 \pm 0.094$ & $82.63 \pm 0.045$ & $69.73 \pm 0.872$ & $57.71 \pm 0.107$ \\
\texttt{BioBERT} & $85.53 \pm 0.130$ & $81.76 \pm 0.094$ & $78.60 \pm 0.072$ & $57.04 \pm 0.868$ \\
\texttt{MatBERT} & $86.09 \pm 0.170$ & $83.35 \pm 0.085$ & $80.01 \pm 0.027$ & $56.91 \pm 0.434$ \\
\texttt{BatteryBERT} & $86.49 \pm 0.085$ & $82.94 \pm 0.309$ & $78.14 \pm 0.103$ & $59.87 \pm 0.398$ \\
\hline
\texttt{\SB{}\_1} & $85.25 \pm 0.063$ & $80.87 \pm 0.282$ & $82.75 \pm 0.049$ & $55.34 \pm 0.742$ \\
\texttt{\SB{}\_10} & $85.80 \pm 0.094$ & $80.61 \pm 0.747$ & $83.24 \pm 0.063$ & $53.41 \pm 0.380$ \\
\texttt{\SB{}\_100} & $85.90 \pm 0.063$ & $82.09 \pm 0.022$ & $83.12 \pm 0.085$ & $54.93 \pm 0.063$ \\
\texttt{\SB{}\_10\_WB} & $83.94 \pm 0.058$ & $78.98 \pm 1.190$ & $83.00 \pm 0.250$ & $54.29 \pm 0.080$ \\
\texttt{\SB{}\_100\_WB} & $84.31 \pm 0.080$ & $80.84 \pm 0.161$ & $82.43 \pm 0.031$ & $54.00 \pm 0.425$ \\
\texttt{\SB{}-XL\_1} & $85.81 \pm 0.054$ & $82.84 \pm 0.228$ & $81.09 \pm 0.170$ & $55.94 \pm 0.899$ \\
\texttt{\SB{}-XL\_100} & $85.73 \pm 0.058$ & $81.75 \pm 0.367$ & $80.72 \pm 0.174$ & $54.54 \pm 0.389$ \\
\hline
\end{tabular}
\end{small}

\vspace{-1ex}

\caption{NER F1 scores for each of 14 models (rows), when the model is finetuned on eight different domain datasets and the resulting finetuned model applied to that dataset's associated NER task (columns). In each case, we give the average value and its standard deviation over five runs.}
\label{tab:token-cls-appendix}
\end{table*}

\begin{table*}[ht!]
\centering
\begin{small}
\begin{tabular}{|l|cccc|}
\hline
{} & \textbf{SciERC} & \textbf{ChemProt} & \textbf{PaperField} & \textbf{Battery} \\
\hline\hline
\texttt{BERT-Base} & $74.95 \pm 1.596$ & $83.70 \pm 0.472$ & $72.83 \pm 0.082$ & $96.31 \pm 0.087$ \\
\texttt{BERT-Large} & $80.14 \pm 2.266$ & $88.06 \pm 0.353$ & $73.12 \pm 0.125$ & $96.90 \pm 0.156$ \\
\texttt{SciBERT} & $79.26 \pm 0.498$ & $89.80 \pm 0.263$ & $73.19 \pm 0.046$ & $96.38 \pm 0.153$ \\
\texttt{PubMedBERT} & $77.45 \pm 0.964$ & $91.78 \pm 0.096$ & $73.93 \pm 0.099$ & $96.58 \pm 0.148$ \\
\texttt{BioBERT} & $80.12 \pm 0.179$ & $89.27 \pm 0.281$ & $73.07 \pm 0.074$ & $96.06 \pm 0.200$ \\
\texttt{MatBERT} & $79.85 \pm 0.121$ & $88.15 \pm 0.026$ & $71.50 \pm 0.135$ & $96.33 \pm 0.106$ \\
\texttt{BatteryBERT} & $78.14 \pm 0.550$ & $88.33 \pm 0.939$ & $73.28 \pm 0.022$ & $96.06 \pm 0.437$ \\
\hline
\texttt{\SB{}\_1} & $73.01 \pm 0.248$ & $83.04 \pm 0.150$ & $72.77 \pm 0.060$ & $94.67 \pm 0.671$ \\
\texttt{\SB{}\_10} & $75.95 \pm 0.203$ & $82.92 \pm 0.792$ & $72.94 \pm 0.182$ & $92.83 \pm 3.758$ \\
\texttt{\SB{}\_100} & $76.19 \pm 1.592$ & $87.60 \pm 0.324$ & $73.14 \pm 0.085$ & $92.38 \pm 5.789$ \\
\texttt{\SB{}\_10\_WB} & $73.17 \pm 1.254$ & $81.48 \pm 1.705$ & $72.37 \pm 0.115$ & $93.15 \pm 1.763$ \\
\texttt{\SB{}\_100\_WB} & $76.71 \pm 2.114$ & $83.98 \pm 0.252$ & $72.29 \pm 0.048$ & $95.55 \pm 0.272$ \\
\texttt{\SB{}-XL\_1} & $74.85 \pm 1.497$ & $90.60 \pm 0.246$ & $73.22 \pm 0.009$ & $88.75 \pm 4.035$ \\
\texttt{\SB{}-XL\_100} & $80.99 \pm 0.900$ & $89.18 \pm 0.499$ & $73.66 \pm 0.113$ & $95.44 \pm 0.100$ \\
\hline
\end{tabular}
\end{small}

\vspace{-1ex}

\caption{
Sentence classification F1 scores for each of 14 models (rows), when the model is finetuned on one of four different domain datasets and the finetuned model is applied to that dataset's associated sentence classification task (columns). In each case, we give the average value and its standard deviation over five runs.}
\label{tab:sent-cls-appendix}
\end{table*}

\end{document}